\documentclass[11pt,a4paper]{article}
\usepackage[hyperref]{acl2021}
\usepackage{times}
\usepackage{latexsym}

\usepackage{microtype}

\aclfinalcopy 
\def\aclpaperid{2637} 

\usepackage{bm}
\usepackage{booktabs}
\usepackage{array}
\usepackage{multirow}
\usepackage{makecell}
\usepackage{arydshln}
\usepackage{graphicx}
\usepackage{color}
\usepackage{amsmath}
\usepackage{amssymb}
\usepackage{tikz}
\usepackage[noend]{algpseudocode}
\usepackage[linesnumbered, ruled, vlined]{algorithm2e}

\title{Learning to Bridge Metric Spaces: Few-shot Joint Learning of Intent Detection and Slot Filling}

\author{
	Yutai Hou\footnotemark[1]~,
	Yongkui Lai\footnotemark[1]~,
	Cheng Chen,
	Wanxiang Che\footnotemark[2]~,
	Ting Liu
	\\
	Research Center for Social Computing and Information Retrieval, \\ Harbin Institute of Technology \\
	{ \{ythou, yklai, cchen, car, tliu\}@ir.hit.edu.cn}  \\
 \\
}

\date{}

\begin{document}
\maketitle
\renewcommand{\thefootnote}{\fnsymbol{footnote}}
\footnotetext[1]{Equal contributions.}
\footnotetext[2]{Corresponding author.}
\renewcommand{\thefootnote}{\arabic{footnote}}
\begin{abstract}

In this paper, we investigate few-shot joint learning for dialogue language understanding.
Most existing few-shot models learn a single task each time with only a few examples.
However, dialogue language understanding contains two closely related tasks, i.e., intent detection and slot filling, and often benefits from jointly learning the two tasks.
This calls for new few-shot learning techniques that are able to capture task relations from only a few examples and jointly learn multiple tasks.
To achieve this, we propose a similarity-based few-shot learning scheme, named \textbf{Con}trastive \textbf{Pro}totype \textbf{M}erging network (\textbf{ConProm}), that learns to bridge metric spaces of intent and slot on data-rich domains, and then adapt the bridged metric space to specific few-shot domain.
Experiments on two public datasets, Snips and FewJoint, show that our model significantly outperforms the strong baselines in one and five shots settings.

\end{abstract}

\section{Introduction}
Few-Shot Learning (FSL) that committed to learning new problems with only a few examples \cite{miller2000learning,matching}
is promising to break the data-shackles of current deep learning.
Commonly, existing FSL methods learn a single few-shot task each time.
But, real-world applications, such as dialogue language understanding, usually contain multiple closely related tasks (e.g., intent detection and slot filling) and often benefit from jointly learning these tasks \cite{worsham2020multi,chen2019bert,qin2019stack,goo2018slot}.
In few-shot scenarios, such requirements of joint learning present new challenges for FSL techniques to capture task relations from only a few examples and jointly learn multiple tasks. 

\begin{figure}[t]
	\centering
	\footnotesize
	\includegraphics[scale=0.4]{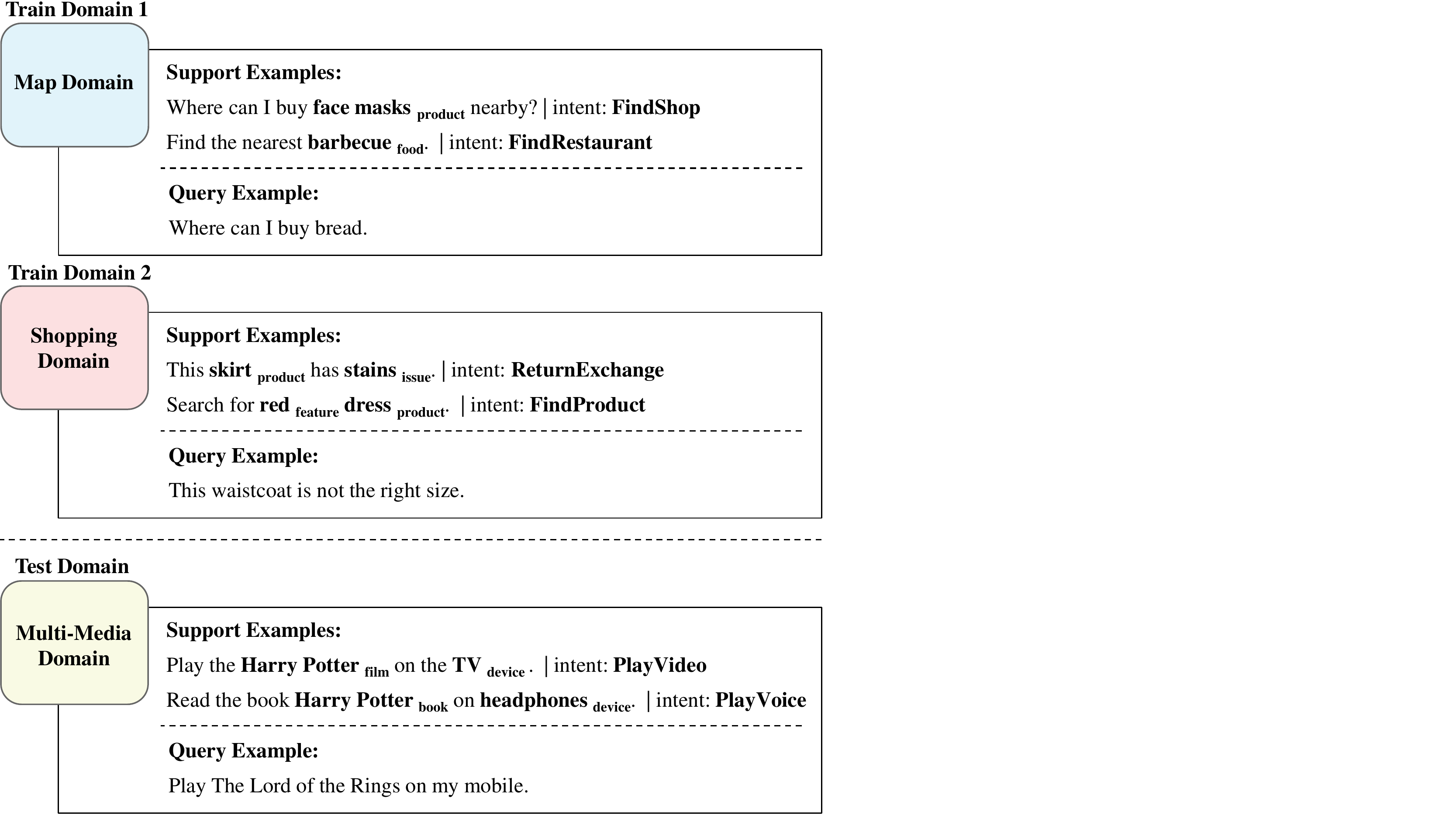}
	\caption{
		Examples of the few-shot joint dialogue language understanding.
		On each domain, given a few labeled support examples, the model predicts the intent and slot labels for unseen query examples.
		Joint learning benefits from capturing the relation between intent and slot labels, but such relation is hard to learn from a few sparse examples and hard to transfer across different domains.
	 }\label{fig:intro}
\end{figure}

This paper explores the few-shot joint learning in dialogue language understanding as an early attempt for this issue.
As shown in Figure \ref{fig:intro}, FSL models are usually first trained on source training domains, then evaluated on an unseen target test domain. Although joint learning can improve dialogue language understanding by utilizing the relation between intents and slots, e.g., ``Harry Potter'' is ``film'' in ``PlayVideo'' intent and ``book'' in ``PlayVoice'' intent, it faces serious challenges when engaging to FSL setting. 
Firstly, it is hard to learn generalized intent-slot relations from only a few support examples. 
Secondly, because the intent-slot relation differs in different domains, it is hard to directly transfer the prior experience from source domains to target domains.
For instance, the intent-slot relation, ``PlayVideo''-``film'', has never appeared in source domains. 


To tackle the aforementioned joint learning challenges in few-shot dialogue language understanding,
we propose the \textbf{Prototype Merging}, which learns the intent-slot relation from data-rich training domains and adaptively captures and utilizes it to an unseen test domain. 
The intent-slot relation is learned with cross-attention between intent and slot class prototypes, which are the mean embeddings of the support examples belonging to the same classes.
Such intent-slot relation adaptively connects the metric spaces of the two tasks. 

Further, to jointly refine the intent and slot metric spaces bridged by Prototype Merging,
we claim that related intents and slots, such as ``PlayVideo'' and ``film'', should be closely distributed in the metric space, otherwise, well-separated.
To achieve this, we propose \textbf{Contrastive Alignment Learning}, which exploits class prototype pairs of related intents and slots as positive samples and non-related pairs as negative samples.
With these samples, it regularizes the FSL process with a margined contrastive loss. 

\begin{figure}[t]
	\centering
	\begin{tikzpicture}
	\draw (0,0 ) node[inner sep=0] {\includegraphics[width=1\columnwidth, trim={0cm 13cm 23.37cm 0cm}, clip]{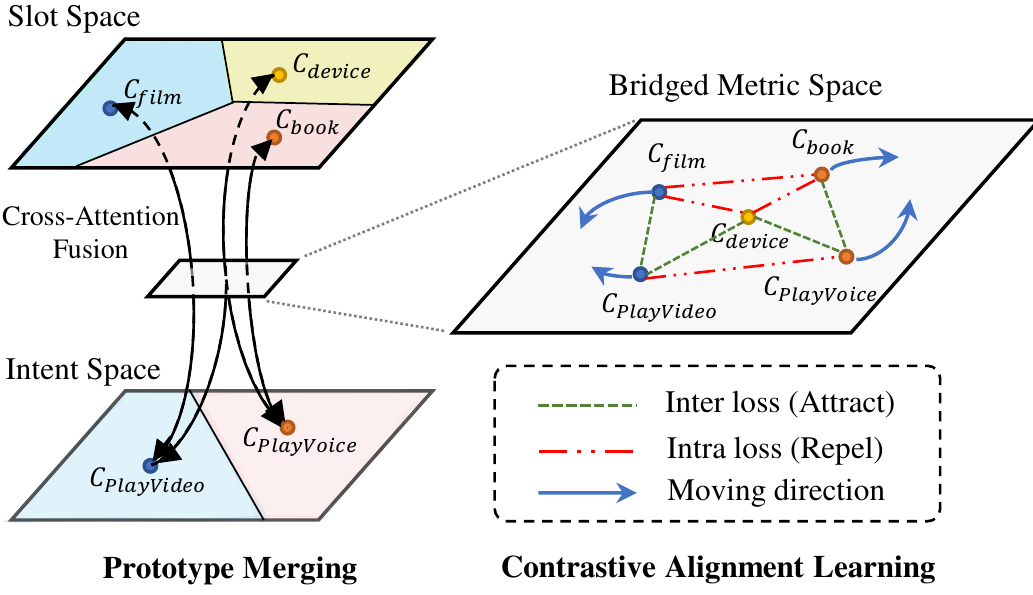}};
	\end{tikzpicture}
	\caption{\footnotesize
		Illustration of two main components of the ConProm model: Prototype Merging and Contrastive Alignment Learning. $C$ denotes prototypes. To ease understanding, we omit the repelling Inter loss in Bridged Metric Space, e.g, loss between $C_\text{book}$ and $C_\text{PlayVideo}$.}\label{fig:model}
\end{figure}

Overall, we named the above novel few-shot joint learning framework as \textbf{Con}trastive \textbf{Pro}totype \textbf{M}erging network (\textbf{ConProm}), which connects intent detection and slot filling tasks by bridging the metric spaces of them.
Two main components of it cooperate to accomplish this goal.
As shown in Figure \ref{fig:model}, Prototype Merging builds the connection between two metric spaces, and Contrastive Alignment Learning refine the bridged metric space by properly distributing prototypes. 


Experiments on two public datasets show both Prototype Merging and Contrastive Aligning Objective significantly boost the few-shot joint learning effects and outperform strong baselines.
In summary, our contribution is three-fold:
(1) We investigate the few-shot joint dialogue language understanding problem, which is also an early attempt for few-shot joint learning problem.
(2) We propose a novel Prototype Merging mechanism to build intent-slot connections adaptively.
(3) We introduce a Contrastive Alignment Learning objective to jointly refines the metric spaces of intent detection and slot filling.
For reproducibility, our code for this paper is publicly available at \url{https://github.com/AtmaHou/FewShotJoint}.

\section{Background}

Before start, we introduce the background of dialogue language understanding and few-shot learning.

\subsection{Dialogue Language Understanding} 
Dialogue language understanding contains two main components: intent detection and slot filling \cite{young2013pomdp}. 
Intent detection is a sentence-level classification problem that classifies a user utterance into one of $N$ intent categories. 

Different from intent detection, slot filling aims to extract key entities within user utterances, which is often achieved by assigning slot tags to each token of a user utterance and is usually formulated as a sequence labeling problem. 
Given input utterance $\bm{x} = \langle x_1, x_2, \ldots, x_n \rangle$ as a sequence of words, joint dialogue language understanding predicts the corresponding semantic frame $\bm{y}=(l, \bm{t})$, 
where $l$ is the intent label and $\bm{t} = \langle t_1, t_2, \ldots, t_n\rangle$ is the slot tags sequence of the utterance.

\subsection{Few-shot Learning}\label{sec:fsl}
Few-shot learning (FSL) extracts prior experience that allows quick adaption to new problems.
Therefore, FSL models are usually first trained on a set of source domains, 
then evaluated on another set of unseen target domains. Figure \ref{fig:intro} shows an example of the training and testing process of few-shot learning for dialogue language understanding.

A target domain only contains a few labeled examples,
which is called support set $\mathcal{S} = \left\{(\bm{x}^{(i)},\bm{y}^{(i)})\right\}_{i=1}^{|\mathcal{S}|}$.
$\mathcal{S}$ includes $K$ examples (K-shot) for each of $N$ classes (N-way).
Taking classification problem as an instance: 
given an input query example $\bm{x} = \langle x_1, x_2, \ldots, x_n \rangle$ and a K-shot support set $\mathcal{S}$ as references, we find the most appropriate class $y^*$ of $\bm{x}$:
\[
	y^*  =  \mathop{\arg\max}_{y}  \ \ p(y  \mid  \bm{x}, \mathcal{S}). 
\]


State-of-the-art few-shot learning is often similarity-based methods \cite{,DBLP:conf/iclr/BaoWCB20,prototypical}.
These methods conquer the extreme lack of data by learning a general similarity metric space on data-rich source domains.
Then on few-shot target domains, they classify a query example according to example-class similarity, where class representations are obtained from a few support examples.

Prototypical network \cite{prototypical} is one of the most classical similarity-based methods.
It obtains the class representation as to the mean embedding of support examples belonging to the same class, so called \textbf{prototypes}:
\[
C_i = \frac{1}{|\mathcal{S}_i|} \sum_{(\bm{x}, y) \in \mathcal{S}_i}{E(\bm{x})}, 
\]
where $\mathcal{S}_i$ is the set of support examples of the $i$th class, and $E(\cdot)$ is the embedding function. 
The probability of $\bm{x}$ belongs to the $i$th class is then made as:
\[
	p(y_i\mid \bm{x}, S) = \frac{\exp{(\textsc{Sim}(E (\bm{x}), C_i)})}{ \sum_j \exp{(\textsc{Sim}(E (\bm{x}), C_j)})}, 
\]
where $\textsc{Sim}(\cdot,\cdot)$ is a vector similarity function.

\section{Proposed Method}
In this section, we introduce the proposed \textbf{Con}trastive \textbf{Pro}totype \textbf{M}erging network (ConProm). 
Firstly, we describe the few-shot intent detection and slot filling with Prototypical network (\S \ref{sec:fslu}).
Based on that, we present two key components of ConProm:
the \textbf{Prototype Merging} mechanism that adaptively connects two metric spaces of intent and slot (\S \ref{sec:pm}) and the \textbf{ Contrastive Alignment Learning} that jointly refines the metric space connected by Prototype Merging (\S \ref{sec:cal}).


\subsection{Few-shot Intent Detection and Slot Filling}\label{sec:fslu}
We build our few-shot intent detection and slot filling model based on the Prototypical Network described in Section \ref{sec:fsl}.
Given a query sentence $\bm{x}$ and a support set $\mathcal{S}$, we estimate the probability of $\bm{x}$ being associated with intent label $l_i$ as:
\begin{flalign} \nonumber
	& p(l_i\mid \bm{x}, \mathcal{S}) \\\nonumber
	& = \frac{\exp{(\textsc{Sim}(E_{\rm intent} (\bm{x}), C_{{\rm intent}_i})})}{ \sum_j \exp{(\textsc{Sim}(E_{\rm intent} (\bm{x}), C_{{\rm intent}_j})})}, 
\end{flalign}
and estimates the probability of the $k$th token in $\bm{x}$ belonging to the $i$th slot class as:
\begin{flalign} \nonumber
& p(t_i\mid k, \bm{x}, \mathcal{S}) \\ \nonumber
& = \frac{\exp{(\textsc{Sim}(E_{\rm slot} (x_k), C_{{\rm slot}_i})})}{ \sum_j \exp{(\textsc{Sim}(E_{\rm slot} (x_k), C_{{\rm slot}_j})})},
\end{flalign}
where $C_{{\rm intent}_i}$ and $C_{{\rm slot}_i}$ are prototypes derived with support examples.
$E_{\rm intent}(\cdot)$ and $E_{\rm slot}(\cdot)$ are embedder functions for intent and slot respectively. 
We adopt BERT \cite{BERT} as the embedder, and the sentence embedding $E_{\rm intent}(\bm{x})$ is calculated as the averaged embedding of its tokens.
We use the dot-product similarity for function $\textsc{Sim}(\cdot, \cdot)$.

\subsection{Prototype Merging}\label{sec:pm}
\begin{figure}[t]
	\centering
	\begin{tikzpicture}
	\draw (0,0 ) node[inner sep=0] {\includegraphics[width=1\columnwidth, trim={0cm 14.6cm 23.5cm 0cm}, clip]{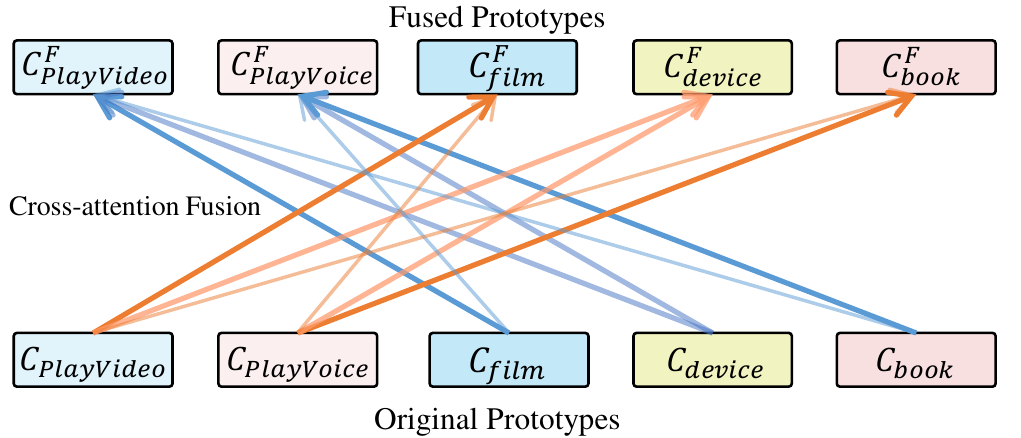}};
	\end{tikzpicture}
	\caption{\footnotesize
		Illustration of cross-attention based information fusion in Prototype Merge. Thicker lines indicate higher cross-attention scores. For example, ``PlayVideo'' and ``film'' are more related, so the corresponding score is larger.}\label{fig:pm}
\end{figure}
To achieve few-shot joint learning and capture the intent-slot relation with the similarity-based method described above, 
we need to bridge the metric spaces of intent detection and slot filling.
However, as mentioned in the introduction, intent-slot relation differs in different domains, it is hard to transfer the bridged metric space learned from source domains to target domains.

To remedy this, we propose the \textbf{Prototype Merging} that can bridge metric spaces adaptively.
As shown in Figure \ref{fig:pm}, Prototype Merging adaptively estimates intent-slot relevance with cross-attention between intent and slot, and then merges the intent and slot prototypes with attentive information fusion.
Such an attentive fusion process enables both intent and slot prototype representations to reflect intent-slot relation and improves domain transferability.


On an unseen target domain, we estimate the intent-slot cross-attention scores from the support set with two methods: (1) use the statistic of co-occurrence of different intents and slots; (2) estimate the intent-slot relevance score using prototype representations.

Firstly, for the statistic-based attention-score, we estimate intent-slot attention scores $A^{\rm S}$ by counting the co-occurrence of different intents and slots, where $A^{\rm S}_{i,j}$ records the normalized number of co-occurrence times for the $i$th intent and the $j$th slot (normalized by row). 

Secondly, for representation-based attention-score, we estimate the cross-attention scores with the Additive Attention \cite{bahdanau2014neural}:\footnote{We adopt additive attention because we find it outperforms common product-based attention in our setting. This is mainly due to that additive attention interferes less with product-based similarity calculations.}
\begin{flalign} 
& A^{\rm R}_{i,j} = V^\top {\rm tanh}(W C_{{\rm intent}_i} +  U C_{{\rm slot}_j}), \nonumber
\end{flalign} 
where $A^{\rm R}$ is the attention matrix, and  $A^{\rm R}_{i,j}$ records the cross-attention score between the $i$th intent and the $j$th slot. 
$U$, $V$ and $W$ are parameters learned on source domains,which preserve the general experience of estimating relevance with representations.
$C_{{\rm intent}_i}$ and $C_{{\rm slot}_j}$ are prototypes of $i$th intent and the $j$th slot respectively. 
We normalize $A^{\rm R}$ by row with {\rm softmax} function.

We obtain the final cross-attention score matrix $A$ by combining $A^{\rm S}$ and $A^{\rm R}$.
\begin{flalign} 
& A = \lambda A^{\rm S} + (1 - \lambda) A^{\rm R},  \nonumber
\end{flalign}
where $\lambda$ is the interpolation factor. 

After obtaining the cross-attention scores, we represent each intent by fusing the information of related slot prototypes, where the attention scores are used as fusing weights. Similarly, we use intent prototypes
to represent slots (See Figure \ref{fig:pm}). 
The fusion process is as follows:
\begin{flalign} 
& C_{{\rm intent}_i}^{\rm F} = \sum_{j} A_{ij} \times C_{{\rm slot}_j}, \nonumber\\
& C_{{\rm slot}_j}^{\rm F} = \sum_{i} A_{ij} \times  C_{{\rm intent}_i}, \nonumber
\end{flalign} 
where $C_{{\rm intent}_i}^{\rm F}$ and $C_{{\rm slot}_j}^{\rm F}$ are the fused prototypes of $i$th intent and the $j$th slot respectively.

At last, we obtain the representation of merged prototypes $C'$ by combining the origin prototype $C$ with the fused prototype $C^{\rm F}$:
\begin{flalign} 
 C_{\rm intent}' &= \alpha \times C_{\rm intent}^{\rm F} + (1 - \alpha) \times C_{\rm intent},  \nonumber\\
 C_{\rm slot}' &= \alpha \times C_{\rm slot}^{\rm F} + (1 - \alpha) \times C_{\rm slot}, \nonumber
\end{flalign} 
where the $\alpha$ is a hyper-parameter that controls the importance of intent-slot relation.

\subsection{Contrastive Alignment Learning}\label{sec:cal}
Similarity-based few-shot learning relies heavily on a good metric space, where different classes should be well separated from each other \cite{hou2020fewshot,yoon2019tapnet}. 
In joint-learning scenarios, there are further requests to connect  metric spaces of joint learned tasks and jointly optimize these metric spaces.

In response to the above requests, we argue that the distribution of prototypes of dialogue language understanding should fit these intuitions: (1) different intent prototypes should be far away and the same as slot prototypes (\textit{Intra-Contrastive}); (2) the slot prototypes should close to the related intent prototypes and should be far away from the unrelated intent prototypes (\textit{Inter-Contrastive}).\footnote{A slot is related to an intent means that they used to co-occur in the same semantic frame.}
To achieve these, we introduce a \textbf{Margined Contrastive Loss} to force the model to learn the separation and alignment of intent and slot prototypes.

Firstly, to encourage separation of prototypes from the same task, we regularize the learning of intent and slot prototypes with \textit{Intra-Contrastive} loss $\mathcal{L}_{\rm Intra} = \frac{1}{2} (\mathcal{L}_{\rm Intra-intent} + \mathcal{L}_{\rm Intra-slot})$, where both the $\mathcal{L}_{\rm Intra-intent}$ and $\mathcal{L}_{\rm Intra-slot}$ are calculated as: 
\begin{flalign} 
	& \mathcal{L}_{\rm Intra} = \frac{1}{N^2} \sum_{i} \sum_{j} {\rm max}(0, m - \left\| C_i - C_j \right\| )^2, \nonumber
\end{flalign}
where $m$ is the margin value and $N$ is the number of prototypes.
The margin $m$ is important since it can protect metric space from excessive dispersion.

Next, we learn the alignment (separation) between intent prototypes and slot prototypes with \textit{Inter-Contrastive} loss $\mathcal{L}_{\rm Inter}$:
\begin{flalign} 
& \mathcal{L}^\mathcal{R}_i =  \frac{1}{2 |\mathcal{R}_i|} \sum_{j \in {\mathcal{R}_i}} (\left\|C_{{\rm intent}_i} - C_{{\rm slot}_j} \right\|^2), \nonumber \\
& \mathcal{L}^\mathcal{U}_i = \frac{1}{2 |\mathcal{U}_i|} \sum_{k \in {\mathcal{U}_i}} \max(0, m - \left\| C_{{\rm intent}_i} - C_{{\rm slot}_k} \right\|)^2, \nonumber \\
& \mathcal{L}_{\rm Inter} = \sum_{i}^{N_I} (\mathcal{L}^{\rm R}_i + \mathcal{L}^{\rm U}_i), \nonumber
\end{flalign} 
where ${\mathcal{R}_i}$ is the set of slots related to the $i$th intent and ${\mathcal{U}_i}$ is the set of slots that are not related to the $i$th intent. $N_I$ is the number of intents.
Here, we simply obtain the relatedness with the co-occurrence matrix $M^{\rm S}$ in Section \ref{sec:pm}.

Finally, the Margin Contrastive Loss is calculated as:
\begin{flalign}
\mathcal{L}_{\rm Contrastive} = \mathcal{L}_{\rm Inter} + \mathcal{L}_{\rm Intra} \nonumber
\end{flalign}

\subsection{Learning Objective}\label{sec:loss}
In dialogue language understanding task, we joint learn the intent detection task and slot filling by optimizing both losses at the same time.
Specifically, we use CrossEntropy (CE) to calculate the loss for intent detection and slot filling. Combining with the loss of Contrastive Alignment Learning, we train the entire model with the following objective function:
\begin{flalign} \nonumber
	& \mathcal{L}_{all} = {\rm CE}_{\rm intent} + {\rm CE}_{\rm slot} + \mathcal{L}_{\rm Contrastive} \nonumber
\end{flalign}

\section{Experiments}

We evaluate our method on the dialogue language understanding task of 1-shot/5-shot setting, which transfers knowledge from source domains (training) to an unseen target domain (testing) containing only 1-shot/5-shot support set.

\subsection{Settings}
\paragraph{Dataset} 
We conduct experiments on two public datasets: Snips \cite{Snips} and FewJoint \cite{FewJoint}. Snips is a widely-used dataset for dialogue language understanding, containing seven single-intent domains together with 53 slots. 
The other dataset FewJoint is joint dialogue language understanding used in the few-shot learning contest of
SMP2020-ECDT Task-1.\footnote{The Eighth China National Conference on Social Media Processing \url{https://smp2020.aconf.cn/smp.html}}
It contains 59 multi-intent domains, 143 different intents, and 205 different slots.

In the few-shot learning setting, we train models on several source domains and test them on unseen target few-shot domains. 
For Snips, we follow \citet{krone2020handful} and combine single-intent domain into multi-intent domain to achieve the classification of intents.
After that, we split the Snips dataset into 3 parts:  the training domain with 3 intents, the developing domain with 2 intents and the testing domain with 2 intents. 
FewJoint is already a few-shot learning benchmark. Therefore, we follow the original data split and there are 45 domains for training, 5 domains for developing and 9 domains for testing.

\paragraph{Few-shot Dataset Construction}
To simulate the few-shot learning situation, we follow previous few-shot learning works \cite{matching,krone2020handful,finn2017maml} and construct the dataset into a few-shot episode style, where the model is trained and evaluated with a series of few-shot episodes.
Each episode contains a support set and query set. 
However, different from the single-task problem,  joint-learning examples are associated with multiple labels. 
Therefore, we cannot guarantee that each label appears $K$ times while sampling examples for the $K$-shot support set.
To remedy this, we build support sets with the Mini-Including Algorithm \cite{hou2020fewshot}, which is intended for such situations.
It constructs support set generally following two criteria: (1) All labels appear at least $K$ times in support set. (2) At least one label will appear less than $K$ times in the support set if any support example is removed from the support set.
For Snips, we construct 200 few-shot episodes for training, 50 for developing, and 50 for testing. 
We set the query set size as 16 for training and developing,  100 for testing.
For FewJoint, we use the few-shot episodes provided by the original dataset. 

\paragraph{Evaluation} 

We adopt three metrics for evaluation: Intent Accuracy, Slot F1-score, Joint Accuracy.\footnote{We calculate the Slot F1-score with the conlleval script \url{https://www.clips.uantwerpen.be/conll2000/chunking/conlleval.txt}}
For joint dialogue language understanding, Joint Accuracy is the most important metric among all three metrics \cite{FewJoint}.
It evaluates the sentence level accuracy, which considers one sentence is correct only when all its slots and intents are correct.

To conduct a robust evaluation under few-shot setting, we validate the models on multiple few-shot episodes (i.e., support-query set pairs) from different domains and take the average score as final results.
To control the non-deterministic neural network training \citep{reimers-gurevych:2017:EMNLP2017}, 
we report the average score of 5 random seeds for all results.

\begin{table*}[t]
	\centering
	\footnotesize
	\renewcommand\arraystretch{1.2}
		\begin{tabular}{lcrrrrr}
			\toprule
			\multirow{2}{*}{\textbf{Models}} & \multicolumn{3}{c}{\textbf{Snips}} & \multicolumn{3}{c}{\textbf{FewJoint}} \\
			\cmidrule(lr){2-4}
			\cmidrule(lr){5-7}
			& \multicolumn{1}{c}{\textbf{Intent Acc.}} & \multicolumn{1}{c}{\textbf{Slot F1}} & \multicolumn{1}{c}{\textbf{Joint Acc.}} & \multicolumn{1}{c}{\textbf{Intent Acc.}} & \multicolumn{1}{c}{\textbf{Slot F1}} & \multicolumn{1}{c}{\textbf{Joint Acc.}} \\
			\midrule
            \textbf{SepProto} & \textbf{ 98.23\tiny{$\pm 0.66$} } & { 43.90\tiny{$\pm 1.98$} } & { 9.47\tiny{$\pm 2.10$} } & { 66.35\tiny{$\pm 0.51$} } & { 27.24\tiny{$\pm 1.10$} } & { 10.92\tiny{$\pm 0.89$} } \\
			\textbf{JointProto}      & { 92.57\tiny{$\pm 0.57$} } & { 42.63\tiny{$\pm 2.03$} } & { 7.35\tiny{$\pm 1.70$} } & {58.52\tiny{$\pm 0.28$} } & {29.49\tiny{$\pm 1.01$} } & {9.64\tiny{$\pm 0.47$} } \\
			\textbf{LD-Proto} & { 97.25\tiny{$\pm 0.71$} } & { 47.81\tiny{$\pm 2.53$} } & { 10.67\tiny{$\pm 1.99$} } & \textbf{ 67.70\tiny{$\pm 0.65$} } & { 27.73\tiny{$\pm 0.35$} } & { 13.70\tiny{$\pm 0.52$} } \\
			\textbf{LD-Proto+TR} & { 97.53\tiny{$\pm 0.30$} } & {51.03\tiny{$\pm 2.40$} } & { 17.32\tiny{$\pm 2.62$} } & { 67.63\tiny{$\pm 1.42$} } & { 34.06\tiny{$\pm 4.75$} } & { 16.98\tiny{$\pm 2.14$} } \\
			\textbf{ConProm} (Ours) & { 96.67\tiny{$\pm 1.45$} } & { 53.05\tiny{$\pm 0.81$} } & { 21.72\tiny{$\pm 0.97$} } & { 65.26\tiny{$\pm 0.23$} } & { 33.09\tiny{$\pm 1.66$} } & { 16.32\tiny{$\pm 0.75$} }  \\
			\textbf{ConProm+TR} (Ours) & { 96.17\tiny{$\pm 0.76$} } & \textbf{ 55.84\tiny{$\pm 0.85$} } & \textbf{29.72\tiny{$\pm 1.30$} } & { 65.73\tiny{$\pm 0.55$} } & \textbf{ 37.97\tiny{$\pm 0.70$} } & \textbf{ 19.57\tiny{$\pm 1.19$} } \\
			\midrule
			\textbf{JointTransfer}        & { 71.07\tiny{$\pm 4.31$} } & { 38.24\tiny{$\pm 2.19$} } & { 13.28\tiny{$\pm 0.45$} } & { 41.83\tiny{$\pm 2.40$} } & { 26.89\tiny{$\pm 2.72$} } & { 12.27\tiny{$\pm 2.09$} } \\
			\textbf{Meta-JOSFIN}        & { 71.38\tiny{$\pm 0.76$} } & {31.47\tiny{$\pm 0.29$} } & {8.88\tiny{$\pm 0.18$} } & {57.92\tiny{$\pm 0.66$} } & {29.26\tiny{$\pm 0.45$} } & {15.00\tiny{$\pm 0.66$} } \\
			\textbf{LD-Proto+FT} & {83.85\tiny{$\pm 6.21$} } & {45.76\tiny{$\pm 5.24$} } & {17.70\tiny{$\pm 2.67$} } & \textbf{64.70\tiny{$\pm 0.50$} } & {32.15\tiny{$\pm 1.28$} } & {21.32\tiny{$\pm 1.80$} } \\
			\textbf{ConProm+FT} (Ours) & {88.20\tiny{$\pm 3.22$} } & {52.41\tiny{$\pm 2.01$} } & {23.05\tiny{$\pm 1.70$} } & {61.24\tiny{$\pm 0.81$} } & {42.02\tiny{$\pm 0.77$} } & {24.63\tiny{$\pm 1.30$} }  \\
			\textbf{ConProm+FT+TR} (Ours) & \textbf{90.45\tiny{$\pm 0.52$} } & \textbf{56.04\tiny{$\pm 1.75$} } & \textbf{27.80\tiny{$\pm 2.33$} } & {63.67\tiny{$\pm 0.94$} } & \textbf{42.44\tiny{$\pm 0.51$} } & \textbf{27.72\tiny{$\pm 0.95$} }  \\
			\bottomrule
		\end{tabular}
	\caption{
		\footnotesize
		Scores on 1-shot dialogue language understanding task on Snips and FewJoint datasets. \textbf{+FT} denotes finetune model. \textbf{+TR} denotes using the trick of transition rule, which blocks illegal slot prediction, such as ``I'' tag after ``O'' tag. Results above the mid-line are from non-finetune based methods, and results below the mid-line are from finetuning based methods.
	}\label{tbl:1shot}
	
\end{table*}

\begin{table*}[t]
	\centering
	\footnotesize
	\renewcommand\arraystretch{1.2}
		\begin{tabular}{lcrrrrr}
			\toprule
			\multirow{2}{*}{\textbf{Models}} & \multicolumn{3}{c}{\textbf{Snips}} & \multicolumn{3}{c}{\textbf{FewJoint}} \\
			\cmidrule(lr){2-4}
			\cmidrule(lr){5-7}
			& \multicolumn{1}{c}{\textbf{Intent Acc.}} & \multicolumn{1}{c}{\textbf{Slot F1}} & \multicolumn{1}{c}{\textbf{Joint Acc.}} & \multicolumn{1}{c}{\textbf{Intent Acc.}} & \multicolumn{1}{c}{\textbf{Slot F1}} & \multicolumn{1}{c}{\textbf{Joint Acc.}} \\
			\midrule
			
			\textbf{SepProto} & \textbf{99.53\tiny{$\pm 0.11$} } & {53.28\tiny{$\pm 1.85$} } & {14.40\tiny{$\pm 3.00$} } & {75.64\tiny{$\pm 1.51$}} & {36.08\tiny{$\pm 0.65$} }&{15.93\tiny{$\pm 1.85$} }\\
			\textbf{JointProto}      & {99.17\tiny{$\pm 0.09$} } & {50.63\tiny{$\pm 2.01$} } & {13.40\tiny{$\pm 1.44$} } & {70.93\tiny{$\pm 2.45$} }& {39.47\tiny{$\pm 1.05$} } & {14.48\tiny{$\pm 1.11$} } \\
			\textbf{LD-Proto} & {99.40\tiny{$\pm 0.08$} } & {48.96\tiny{$\pm 1.85$} } & {20.93\tiny{$\pm 3.00$} } & \textbf{78.29\tiny{$\pm 1.51$}} & {39.88\tiny{$\pm 0.65$} }&{22.91\tiny{$\pm 1.85$} }\\
			\textbf{LD-Proto+TR} & {99.20\tiny{$\pm 0.30$} } & {54.87\tiny{$\pm 3.79$} } & {29.40\tiny{$\pm 2.90$} } & {75.75\tiny{$\pm 0.95$} } & \textbf{51.62\tiny{$\pm 2.82$} } & {27.59\tiny{$\pm 2.31$} } \\
			\textbf{ConProm} (Ours)& {98.50\tiny{$\pm 0.42$} } & {61.03\tiny{$\pm 1.77$} } & {32.20\tiny{$\pm 2.06$} }  & {78.05\tiny{$\pm 1.04$} } &  {39.40\tiny{$\pm 1.75$} } & {24.18\tiny{$\pm 1.29$} }  \\
			\textbf{ConProm+TR} (Ours) & {98.99\tiny{$\pm 0.14$} } & \textbf{65.13\tiny{$\pm 1.46$} } & \textbf{40.20\tiny{$\pm 2.24$} } & {75.54\tiny{$\pm 1.85$} } & {50.28\tiny{$\pm 1.03$} } & \textbf{28.69\tiny{$\pm 1.61$} } \\
			\midrule
			\textbf{JointTransfer}        & {88.87\tiny{$\pm 5.04$} } & {49.62\tiny{$\pm 1.87$} } & {25.50\tiny{$\pm 3.09$} } & {57.50\tiny{$\pm 6.09$} } & {29.00\tiny{$\pm 4.35$} } & {18.81\tiny{$\pm 4.45$} } \\
			\textbf{Meta-JOSFIN}        & {92.47\tiny{$\pm 1.26$} }  & {56.85\tiny{$\pm 1.25$} } & {25.87\tiny{$\pm 0.31$} } & {78.91\tiny{$\pm 0.53$} } & {53.88\tiny{$\pm 1.63$} } & {36.63\tiny{$\pm 1.01$} } \\
			\textbf{LD-Proto+FT} & {81.07\tiny{$\pm 8.61$} } & {59.27\tiny{$\pm 3.61$} } & {26.33\tiny{$\pm 2.38$} } & \textbf{80.50\tiny{$\pm 0.97$} } &{55.33\tiny{$\pm 2.55$} }  & {38.11\tiny{$\pm 2.60$} } \\
			\textbf{ConProm+FT} (Ours) & {96.23\tiny{$\pm 1.19$} } & {66.66\tiny{$\pm 2.46$} } & {39.87\tiny{$\pm 2.60$} }  & {78.33\tiny{$\pm 1.14$} } & {62.34\tiny{$\pm 0.26$} } & {40.25\tiny{$\pm 1.19$} } \\
			
			\textbf{ConProm+FT+TR} (Ours) & \textbf{98.40\tiny{$\pm 0.20$} } & \textbf{72.98\tiny{$\pm 0.41$} } & \textbf{52.95\tiny{$\pm 0.85$} } & {78.43\tiny{$\pm 1.86$} } & \textbf{69.44\tiny{$\pm 0.39$} } & \textbf{46.54\tiny{$\pm 0.72$} } \\
			\bottomrule
		\end{tabular}
	\caption{
		\footnotesize
		Scores on 5-shot dialogue language understanding task on Snips dataset and FewJoint dataset.
	}\label{tbl:5shot}
	
\end{table*}

\vspace*{-2mm}
\subsection{Baselines}

We compare our model with two kinds of strong baseline: fine-tune based transfer learning methods (JointTransfer, Meta-JOSFIN) and similarity-based FSL methods (SepProto, JointProto, LD-Proto).

\paragraph{JointTransfer} is a domain transfer model based on the JointBERT \cite{chen2019bert}. It consists of a shared BERT embedder with intent detection and slot filling layers on the top. We pretrain it on source domains and finetune it on target domain support sets.

\paragraph{Meta-JOSFIN} \cite{bhathiya2020metajoint} is a meta-learning model based on the MAML \cite{finn2017maml}.
The meta-learner model here is a BERT-based joint dialogue language understanding model similar to \textbf{JointTransfer}.
It learns initial parameters that can fast adapt to the target domain after only a few updates.

\paragraph{SepProto} is a prototypical-based dialogue language understanding model with BERT embedding, that learns intent detection and slot filling separately.
During the experiment, it is pre-trained on source domains and then directly applies to target domains without fine-tuning.

\paragraph{JointProto} \cite{krone2020handful} is all the same as SepProto except that it jointly learns the intent and slot tasks by sharing the BERT encoder. 




\paragraph{LD-Proto} is also a prototypical model similar to \textbf{JointProto}. 
The only difference is that it is enhanced by the logits-dependency tricks \cite{goo2018slot}, where joint learning is achieved by depending on the intent and slot prediction on the logits of the accompanying task.

\paragraph{Implements}
For both ours and baseline models, we determine the hyperparameters on the development set.
We use ADAM \citep{DBLP:journals/corr/KingmaB14} for training and set batch size as 4 and learning rate as $10^{-5}$.
We adopt embedding tricks of Pairs-Wise Embedding \cite{relationGao2019fewrel,hou2020fewshot} and Gradual Unfreezing \cite{Howard2018UniversalLM}.
The $\lambda$ and $\alpha$ in Section \ref{sec:pm} are both set as 0.5.
We implement both our and baseline models with the few-shot platform MetaDialog.\footnote{\url{https://github.com/AtmaHou/MetaDialog}}
Besides, to use the information in target domains and make a fair comparison with fine-tuning baselines, we explore the performance of the similarity-based model under fine-tuning setting (+FT) and enhance the model with a fine-tune process similar to Meta-JOSFIN.
In addition, following the suggestions of \citet{hou2020fewshot}, we investigate adding Transition Rules (+TR) between slot tags, which bans illegal slot prediction, such as ``I'' tag after ``O'' tag.

\vspace*{-1mm}
\subsection{Main Results}\label{sec:main_res}
In this section, we present the evaluation of the proposed method on both 1-shot and 5-shot dialogue understanding setting.

\vspace*{-1mm}
\paragraph{Result of 1-shot setting} 
As shown in Table \ref{tbl:1shot}, our method (ConProm) achieves the best performance on Joint Accuracy, which is the most important metric.
Among all metrics, ConProm only lags a bit than LD-Proto on intent accuracy.
We address this to the fact that there are many slots shared by different intent, and representing an intent with slots may unavoidably introduce noise from other intents. 
Considering the huge improvements on Slot and Joint performance over LD-Proto, we argue that the limited loss is a worthy compromise here.
Since similarity-based models predict slot tags independently for each token, they tend to predict illegal tags.
We employ a simple transition rule (+TR) to remedy such defects and further improves the performance.
For fairness, we also enhance LD-Proto with TR trick and our model still outperforms the enhanced baseline.

For those non-finetuned methods,
ConProm outperforms LD-Proto by Joint Accuracy scores of 11.05 on Snips and 2.62 on FewJoint, which show that our model can better capture the relation between intent and slot.
Our improvements on Snips are higher than those on FewJoint, which is mainly because that there is clearer intent-slot dependency in Snips. 
The performance of JointProto is even lower than SepProto, which demonstrates that few-shot joint learning is not a trivial issue as simply sharing the embeddings

When finetuning brings significant improvements for all methods, our model (ConProm+FT) still achieves the best performance. 
Interestingly, we observe that finetuning often hurts the intent prediction. 
This shows that finetuning brings limited gains on sentence-level domain knowledge but leads to overfitting.


\vspace*{-1mm}
\paragraph{Result of 5-shot setting} Table \ref{tbl:5shot} shows the 5-shot results. The results are consistent with 1-shot setting in general trending and our methods achieve the best performance.
While more learning shots improve the performance for all methods, the superiority of our best performed baseline is further strengthened.
This shows that the model can better exploit the richer intent-slot relations hidden in 5-shot support sets.

\vspace*{-1mm}

\subsection{Analysis}

\begin{table}[t]
	\centering
	\footnotesize
	\renewcommand\arraystretch{1.2}
		\begin{tabular}{lcccc}
			\toprule
			\multirow{2}{*}{\textbf{Setting}} & \multicolumn{2}{c}{\textbf{Snips}} & \multicolumn{2}{c}{\textbf{FewJoint}} \\
			\cmidrule(lr){2-3}
			\cmidrule(lr){4-5}
			& \multicolumn{1}{c}{\textbf{1-shot}} & \multicolumn{1}{c}{\textbf{5-shot}} & \multicolumn{1}{c}{\textbf{1-shot}} & \multicolumn{1}{c}{\textbf{5-shot}} \\
			\midrule
			Ours & { 21.72 } & { 32.20 } & { 16.32 } & { 24.18 }\\
			- PM & { -1.90 } & { -2.63 } & { -4.90 } & { -8.39 }\\
			- CAL & { -5.19 } & { -12.73 } & { -1.78 } & { -3.78 }\\
			\bottomrule
		\end{tabular}
	\caption{
		\footnotesize
		Ablation study over two main components of proposed framework: Prototype Merge (PM) and Contrastive Alignment Learning (CAL). The score is Joint Accuracy.
	}\label{tbl:abl}
		\vspace*{-2mm}
	
\end{table}

\paragraph{Ablation Test}  To inspect how each component of the proposed model contributes to the final performance, we conduct ablation analysis. 
As shown in Table \ref{tbl:abl}, we independently removing two main components: Prototype Merge (PM) and Contrastive Alignment Learning (CAL). 


When PM is removed, the intent and slot prototypes are represented only with corresponding support examples, and Joint Accuracy drops are witnessed. 
There is more loss on FewJoint. Because there are much more slots shared by different intents in FewJoint, and the attention mechanism of PM is important for identifying relatedness between intents and slots.

For our model without CAL, we train the model with only cross entropy loss and get lower scores on all settings.
There are more performance drops on Snips.
This is mainly because that there much clearer intent-slot relation in Snips, which can be easily handled by CAL. 

In terms of contribution, there are opposite performance for CAL and PM on two dataset, which shows that PM and CAL complement each other and reach a balance for various situations.

\begin{figure}[t]
	\centering
	\begin{tikzpicture}
	\draw (0,0 ) node[inner sep=0] {\includegraphics[width=1\columnwidth, trim={2cm 5.1cm 1.6cm 0.5cm}, clip]{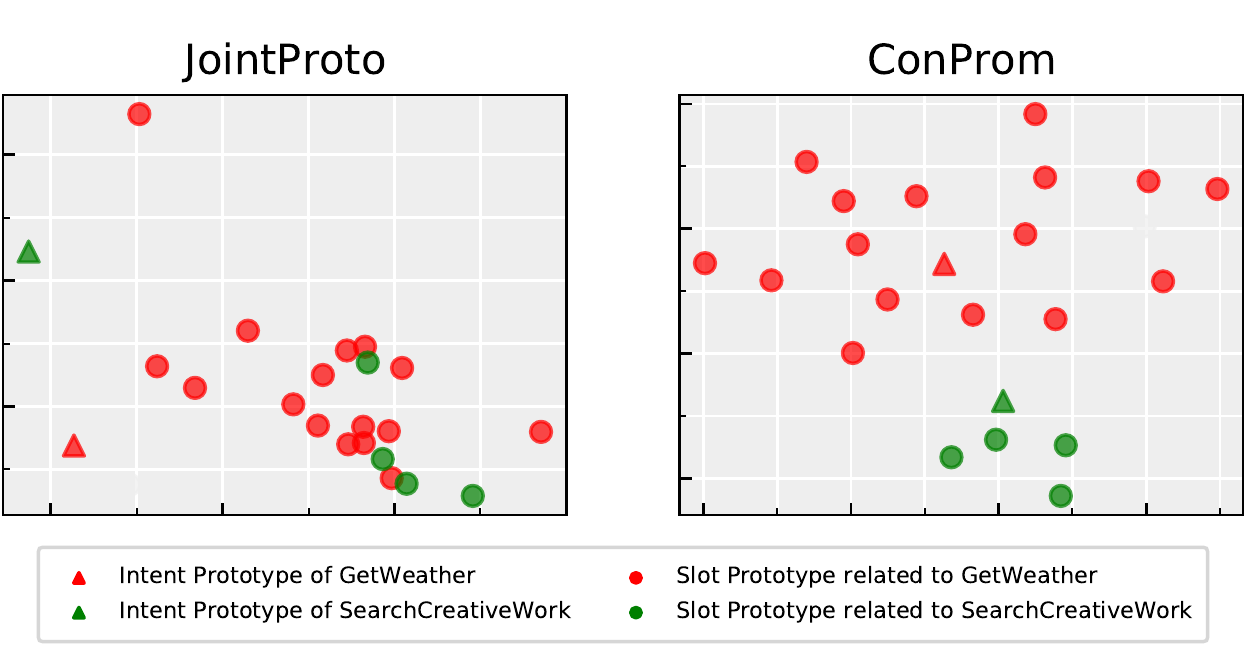}};
	\end{tikzpicture}
	\caption{\footnotesize
		Visualization of the prototype distribution of JointProto and Ours (ConProm) with tSNE (step=500). 
	 }\label{fig:visual}
\end{figure}

\vspace*{-1mm}
\paragraph{Visual Analysis of Prototype Distribution} 
To get further an understanding of the model effects on bridging the metric spaces of intent and slot, 
we visualize the prototype distributions in the metric space.
As shown in Figure \ref{fig:visual}, it is exciting to see that our model successfully refine the prototype distribution by aligning the slots to related intent and making prototypes properly well-separated. 

\vspace*{-1mm}
\paragraph{Sentence level slot accuracy analysis} There is some confusion in Table \ref{tbl:1shot} and Table \ref{tbl:5shot} that there are huge performance differences of Joint Accuracy score when Intent Accuracy scores and Slot F1 scores are similar. 
We inspect this issue by evaluating the Sentence Level Slot Accuracy, which considers a sentence to be correct when all slots are correct. As shown in Table \ref{tbl:slot_acc_1shot}, there is a huge gap in the slot accuracy score between LD-Proto and ConProm, which explains the gap in Joint score.

\begin{table}[t]
	\centering
	\footnotesize
	\renewcommand\arraystretch{1.2}
		\begin{tabular}{lrrrrrrrrrrr}
			\toprule
			\multirow{2}{*}{\textbf{Models}} & \multicolumn{2}{c}{\textbf{Snips}} & \multicolumn{2}{c}{\textbf{FewJoint}} \\
			\cmidrule(lr){2-3}
			\cmidrule(lr){4-5}
			& \multicolumn{1}{c}{\textbf{F1.}} & \multicolumn{1}{c}{\textbf{Acc.}} & \multicolumn{1}{c}{\textbf{F1.}} & \multicolumn{1}{c}{\textbf{Acc.}} \\
			\midrule
			\textbf{JointProto}      & 42.63 & 8.08 & 29.49 & 15.73 \\
			\textbf{LD-Proto} & 47.81 & 10.72 & 27.73 & 20.44 \\
			\textbf{LD-Proto+TD} & 51.03 & 17.53 & 34.06 & 24.69 \\
			\textbf{ConProm} & 53.05 & 22.30 & 33.09 & 22.38 \\
			\textbf{ConProm+TD} & 55.84 & 30.47 & 37.97 & 26.31 \\
			\midrule
			\textbf{JointTransfer}        & 38.24 & 14.38 & 26.89 & 26.37 \\
			\textbf{Meta-JOSFIN}        & 31.47 & 9.73 & 29.26 & 21.73 \\
			\textbf{LD-Proto+FT} & 45.76 & 21.92 & 32.15 & 35.75 \\
			\textbf{ConProm+FT} & 52.41 & 25.97 & 42.02 & 39.62 \\
			\textbf{ConProm+TD+FT} & 56.04 & 27.91 & 42.44 & 40.71 \\
			\bottomrule
		\end{tabular}
	\caption{
		\footnotesize
	    Analysis for sentence level slot accuracy.
	    \vspace*{-1mm}
	}\label{tbl:slot_acc_1shot}
	
\end{table}

\section{Related Work}
Few-shot learning is one of the most important direction for machine learning area \cite{fei2006knowledge, fink2005object} and often achieved by similarity-based method \cite{matching} and fine-tuning based method \cite{finn2017maml}. 
FSL in natural language processing has been explored for various tasks, 
including text classification \cite{textSun2019hierarchical,textGeng2019induction,yan2018few,yu2018diverse}, 
entity relation classification 
\cite{relationLv2019adapting,relationGao2019neural,relationYeL19}, sequence labeling \cite{slotLuo2018marrying,hou2018sequence,Shah2019robust,hou2020fewshot,liu2020coach}.

As the important part of a dialog system, dialogue language understanding attract a lot of attention in few-shot scenario. 
\citet{dopierre2020shot,policyVlasov2018few,xia2018zero} explored few-shot intent detection technique.
\citet{slotLuo2018marrying} and \citet{hou2020fewshot} investigated few-shot slot tagging by using prototypical network. 
\citet{hou2021mlc} explored few-shot multi-label intent detection with an adaptive logit adapting threshold. 
But all of these works focus on a single task.

Despite a lot of works on joint dialogue understanding \cite{goo2018slot,li2018self,zhang2018joint,qin2019stack,wang2018bi,niu2019novel,wu2020slotrefine,gangadharaiah2019joint,liu2019cm,qin2020towards}, few-shot joint dialogue understanding is less investigated.
\citet{krone2020learning} and \citet{bhathiya2020metajoint} make the earliest attempts by directly adopt general and classic few-shot learning methods such as MAML and prototypical network.
These methods achieve joint learning by sharing the embedding between intent detection and slot filling task, which model the relation between intent and slot task implicitly.
By contrast, we explicitly model the interaction between intent and slot with attentive information fusion and constrastive loss.
Experiment results also demonstrate the superiority of our method on this task. 

\vspace*{-1mm}

\section{Conclusion}


In this paper, we propose a similarity-based few-shot joint learning framework, ConProm, for dialogue understanding. 
To adaptively model the interaction between intents and slots, we propose the Prototype Merging that bridges the intent metric and slot metric spaces with cross-attention between intent and slot.
To learn better bridged metric space for intent and slot, we propose the Contrastive Alignment Learning to align related cross-task labels in metric space and force unrelated labels properly separated.
Experiment results validate that both Prototype Merging and Contrastive Alignment Learning can improve performance.

\section*{Acknowledgments}
We are grateful for the helpful comments and suggestions from the anonymous reviewers. 
This work was supported by the National Key R\&D Program of China via grant 2020AAA0106501 and the National Natural Science Foundation of China (NSFC) via grant 61976072 and 61772153.

\bibliographystyle{acl_natbib}
\bibliography{acl2021_new_concise}


\end{document}